\documentclass[preprint,authoryear,11pt,nopreprintline]{elsarticle}

\usepackage[letterpaper,margin=1in]{geometry}
\ifdefined\pdfminorversion\pdfminorversion=7\fi
\usepackage[utf8]{inputenc}
\usepackage[T1]{fontenc}
\usepackage{lmodern}
\usepackage{graphicx}
\usepackage{booktabs}
\usepackage{xcolor}
\usepackage{url}
\usepackage{microtype}
\usepackage{rotating}

\usepackage{amsmath,amsfonts,amsthm,bm}
\usepackage{yhmath}
\usepackage{derivative}
\usepackage{algorithm}
\usepackage{algorithmic}

\newcommand{\mask}{\text{\texttt{[MASK]}}}
\newcommand{\pad}{\text{\texttt{[PAD]}}}
\newcommand{\missing}{\text{\texttt{[MISSING]}}}
\newcommand{\na}{\text{\texttt{[NA]}}}

\newcommand{\pdata}{p_{\rm{data}}}

\def\sE{{\mathbb{E}}}

\def\vtheta{{\bm{\theta}}}

\usepackage[hidelinks]{hyperref}
\usepackage{etoolbox}
\makeatletter
\patchcmd{\pprintMaketitle}
  {\Large\@title\par\vskip18pt}
  {{\fontsize{17}{20}\selectfont\bfseries\@title\par}\vskip16pt}
  {}{\PackageWarning{arxiv-layout}{Title styling patch not applied}}
\patchcmd{\pprintMaketitle}
  {\footnotesize\itshape\elsaddress\par\vskip36pt}
  {\footnotesize\upshape\elsaddress\par\vskip20pt}
  {}{\PackageWarning{arxiv-layout}{Affiliation styling patch not applied}}

\patchcmd{\abstract}
  {\noindent\unskip\textbf{\@elsarticleabstitle}}
  {\noindent\makebox[\textwidth][c]{\textbf{\@elsarticleabstitle}}}
  {}{\PackageWarning{arxiv-layout}{Abstract centering patch not applied}}
\patchcmd{\pprintMaketitle}
  {\hrule\vskip12pt}{\vskip12pt}
  {}{\PackageWarning{arxiv-layout}{Upper abstract rule patch not applied}}
\patchcmd{\pprintMaketitle}
  {\hrule\vskip12pt}{\vskip12pt}
  {}{\PackageWarning{arxiv-layout}{Lower abstract rule patch not applied}}
\makeatother

\journal{Transportation Research Part C: Emerging Technologies}

\begin{document}

\begin{frontmatter}

\title{Masked Diffusion for Travel Diaries Synthesis}

\author[aff1]{Pascal Jutras-Dub\'e\corref{cor1}}
\ead{pjutrasd@purdue.edu}
\author[aff2]{Raziq Zaman}
\author[aff3]{Fabian Bastin}
\author[aff2]{Cinzia Cirillo}
\author[aff1]{Ruqi Zhang}

\cortext[cor1]{Corresponding author.} 

\affiliation[aff1]{organization={Department of Computer Science, Purdue University},city={West Lafayette},country={United States}}
\affiliation[aff2]{organization={Department of Civil and Environmental Engineering, University of Maryland},city={College Park},country={United States}}
\affiliation[aff3]{organization={Department of Computer Science and Operations Research, Université de Montréal},city={Montréal},country={Canada}}

\begin{abstract}
Because household travel surveys combine household characteristics, variable-length person rosters, vehicle rosters, and trip chains, existing synthesis pipelines split them across separate models or flatten them into cropped fixed-length vectors. Representation, rather than modeling, becomes the bottleneck: splitting severs cross-segment dependence, and cropping truncates the household-structure distribution. We instead tokenize the diary and keep the object whole: each household-day becomes one token sequence with count tokens for vehicles, persons, and trips, and distinct symbols for padded, hidden, unobserved, and inapplicable positions. We introduce \emph{DiaryBridge}, a masked diffusion model that \emph{bridges} the fully hidden sequence to the distribution of complete \emph{diaries} by unmasking positions in any order. One trained model, conditioned on any observed subset, performs synthesis and partial-diary completion alike. Experiments on real data extracted from the 2017/2018 Regional Travel Survey of the Baltimore region show that DiaryBridge generates travel reflecting vehicle availability and employment status, recovers shared trips using other household members’ observed diaries, and better reproduces aggregate demand across modes, purposes, and departure periods than the learned baselines. The code is available at this 
\textcolor{blue}{\href{https://github.com/PascalJD/diarybridge-release}{repository}}.
\end{abstract}

\end{frontmatter}

\section{Introduction}
\label{sec:intro}
Household travel diaries are a primary input to activity- and agent-based travel demand models \citep{HORL2021103291,FELBERMAIR2020273}, but they are costly to collect and increasingly affected by declining response rates. Synthetic diaries can complement them: to enlarge or rebalance a sample, to fill in items that respondents left unanswered, or to attach travel to a synthetic population. Downstream uses include demand modeling, equity analysis, and policy testing.
 
Population synthesis in transportation is usually framed as constructing households and persons that are consistent with aggregate controls at one or more geographic levels. Travel diary synthesis is harder: we must also generate the within-household and within-day organization of travel, including who travels, in what order, for what purposes, by which modes, at what times, with which household resources, and with whom. In raw household travel surveys, this object is neither a flat table nor a standard time series. It is a linked relational diary composed of household attributes, person rosters, vehicle ownership, and ordered person-day trip chains.
 
Most current pipelines still decompose the problem. Classical fitting, reweighting, and microsimulation methods can match household and person marginals, but they do not natively generate full daily diaries. Sequence models for activities and tours capture temporal order, but often leave household coupling and vehicle information outside the core model. Even when recent deep generative methods are used, they often flatten the diary into one fixed vector with aggressive caps, making representation, rather than modeling, the main bottleneck.
 
Our starting point is therefore to separate the \emph{raw diary object} from the \emph{model representation}. We introduce a canonical encoding that writes each household-day as a single fixed-length sequence of \emph{tokens}, where a token is one discrete value at one position of the record, for example the household’s income category or the mode of a person’s third trip. Explicit count tokens are introduced to record how many vehicles, persons, and per-person trips the household realizes. Because households differ in size, the sequence reserves room for a maximum number of vehicles, persons, and trips per person, and padding fills the slots that a household does not use, much as unused rows of a paper survey roster are left blank. Slots are padded rather than cropped away, so every household within these caps is represented exactly. Padding is not conflated with other kinds of absence: in survey terms, four distinct symbols separate unused roster capacity, values hidden for the model to generate, item nonresponse, and questions that do not apply.
 
This choice makes the modeling problem concrete. We learn \emph{DiaryBridge}, a masked discrete diffusion model over the fixed-length structured encoding. During training, tokens are hidden (masked) at random and the model learns to restore them from the visible ones; at generation time, every token to be generated starts hidden and is revealed progressively. Formally, let $X_{\mathrm{obs}}$ denote any observed subset of tokens and $X_{\mathrm{mis}}$ denote the complement. The target is a single conditional sampler
\[
    p_{\vtheta}(X_{\mathrm{mis}} \mid X_{\mathrm{obs}}),
\]
which reduces to unconditional generation when $X_{\mathrm{obs}}=\emptyset$ or to diary completion when arbitrary fields are empty.
 
Two design decisions distinguish this formulation from applying an off-the-shelf generative model to a flattened table. First, the encoding mixes vocabularies: a token at a household-income position and a token at a departure-time position live in different value spaces. The encoding therefore factorizes the vocabulary by position, and legality constraints are enforced in both the training objective and the sampler. Second, the count tokens and the padding pattern are logically coupled: a sample that declares two household members must pad every person slot from the third onward. We resolve this coupling with a two-phase \emph{count-first sampling} procedure that exploits the fact that masked diffusion can reveal positions in any order: the counts are generated first, the padding pattern implied by the counts is then written deterministically, and only the realized slots are generated afterwards. Generated diaries are therefore structurally consistent by construction, without a separate structure model and without post-hoc repair.
 
The contributions of the paper are:
\begin{enumerate}
    \item A canonical diary encoding that preserves household, vehicle, person, and ordered-trip structure, with explicit count tokens and a four-way distinction between padded, hidden, unobserved, and inapplicable positions, while remaining compatible with standard masked generative models.
    \item A masked discrete diffusion model over this encoding with position-factorized vocabularies enforced during training and sampling, and a count-first sampling procedure that guarantees the structural consistency of every generated diary.
    \item A travel-focused evaluation of household-day synthesis and completion, examining whether modes and purposes reflect vehicle availability and employment status, whether other members’ diaries improve shared-trip recovery, and whether generated travel reproduces trip volumes and patterns for a given population. Comparisons with survey-reuse methods and learned generators also assess distributional fit, predictive utility, and record validity.
\end{enumerate}
 
The remainder of the paper is organized as follows. Section~\ref{sec:related} reviews related work. Section~\ref{sec:encoding} introduces the canonical diary encoding, and Section~\ref{sec:method} presents the diffusion model, legality masking, and the count-first sampler over it. Section \ref{sec:experiments} describes the 2017/2018 Regional Travel Survey data, the training protocol, and the comparison methods, and Section \ref{sec:results} reports the experiments and their results. Section \ref{sec:conclusion} discusses limitations and concludes.

\section{Related Work}
\label{sec:related}
Synthetic travel diaries serve a different role from synthetic populations alone: they provide the disaggregate daily travel demand required by activity- and agent-based transportation models. Such models represent trips associated with individual travelers, allowing demand to be resolved by time, mode, purpose, and traveler characteristics and subsequently used for network simulation and policy analysis. \citet{HORL2021103291}, for example, construct a synthetic population together with daily activity chains for agent-based simulation of Paris and Île-de-France, while \citet{FELBERMAIR2020273} generate 24-hour activity plans specifically as inputs to agent-based transport models.

Classical transportation population synthesis is usually concerned with constructing household and person records that match aggregate controls, rather than generating full daily travel diaries. Foundational work combined census marginals with survey microdata via iterative proportional fitting and sample allocation to create synthetic baseline populations \citep{Beckman1996Baseline}. Later work extended this logic to jointly satisfy household- and person-level constraints through iterative proportional updating \citep{Ye2009IPU}. To move beyond deterministic cloning and sparse contingency tables, probabilistic synthesis methods introduced MCMC simulation, Bayesian networks, hidden Markov models, and hierarchical mixture models \citep{Farooq2013Simulation,Sun2015BN,Saadi2016HMM,Sun2018Hierarchical}. These methods are central to household and person population synthesis, but their unit of synthesis is typically a static agent table rather than a within-day household travel diary.

A separate literature addresses activity and trip-diary generation more directly. Classical activity-based systems generated daily activity patterns, tours, and schedules through structured behavioral submodels \citep{Bowman2001Activity}. More recent work has used probabilistic graphical models to synthesize daily activity and trip chains \citep{Joubert2020ActivityBN}, and deep generative models have begun to synthesize individual 24-hour activity schedules with variational autoencoders \citep{Shone2025ScheduleVAE}. However, these models generally operate at the person-schedule or trip-chain level. They do not treat the household-day diary as a single linked object spanning household attributes, person rosters, vehicle availability, and ordered trips within each person-day.

Deep generative modeling entered transportation population synthesis through learned models of the joint distribution over agent attributes. Variational autoencoders were adapted to generate high-dimensional mixed tabular micro-agents \citep{Borysov2019MicroAgents}, with follow-up work showing that such models can recover attribute combinations absent from the training sample \citep{Garrido2020RareCombinations}. Composite travel GANs extended generation to agents with both tabular and sequential mobility data \citep{BaduMarfo2022CTGAN}, and later work explicitly studied the trade-off between diversity and feasibility using GAN and VAE losses tied to sampling zeros and structural zeros \citep{Kim2023FeasibleDiverse}. Generative models have also been used for conditional completion and source fusion: transformer-based conditional GANs impute sociodemographic attributes and trip purposes for smart-card trip chains \citep{Kim2022TripChainCGAN}, collaborative GANs fuse household travel survey and smart-card data into heterogeneous activity schedules \citep{Lee2025CollaGAN}, and dedicated fusion methods combine survey and passively collected mobility data \citep{Vo2025DataFusion}. Copula-based approaches address transferability of synthesizers across populations \citep{jutras2024copula}.

Very recently, diffusion models and large language models have entered transportation population synthesis. \citet{Kang2023DiffusionPopSyn} apply a denoising diffusion probabilistic model to person-level attribute tables and report improved standardized error over a VAE baseline. \citet{Tang2025DiffusionPopSyn} propose a diffusion-based synthesizer aimed at balancing feasibility and diversity against VAE and GAN alternatives. \citet{Lim2026LLMPopSyn} fine-tune a large language model whose autoregressive generation order is derived from a Bayesian network, again targeting the feasibility--diversity trade-off. All of these operate on flat person- or agent-level attribute vectors. To our knowledge, no existing transportation work applies \emph{masked discrete} diffusion to the \emph{linked household-day diary} nor formulates synthesis and completion as a single sampler over that object. 

For the tokenized part of our problem, discrete diffusion is especially appealing. Discrete denoising diffusion models introduced structured corruption kernels and absorbing-state masking directly in discrete state spaces \citep{Austin2021D3PM}. Subsequent continuous-time and score-based formulations improved both the tractability and empirical performance of discrete diffusion on large discrete domains \citep{campbell2022ctcm,lou2024sedd}. Recent masked diffusion work further shows that these models can be trained with objectives closely related to masked prediction \citep{ghazvininejad2019maskpredict}, while retaining non-left-to-right generation, bidirectional conditioning, and iterative refinement \citep{sahoo2024mdlm,shi2024md4} with recent hybrid and unifying formulations extending the framework \citep{pynadath2026candi,jutrasdube2026sticky}. This matches our setting more naturally than a fixed autoregressive factorization: in diary completion, the observed information is typically an arbitrary subset of household, person, vehicle, and trip fields rather than a clean left-to-right prefix.

Compared with continuous diffusion in latent or embedding spaces \citep{li2022diffusionlm,dieleman2022cdcd,lovelace2023ld4lg,chen2023analogbits}, masked discrete diffusion remains on the token support and therefore aligns well with a representation in which count tokens delimit realized household structure and where padded, hidden, unobserved, and inapplicable positions carry distinct semantics. This is useful for our canonical diary encoding because inactive slots, intentionally hidden fields, structurally missing values, and inapplicable fields should not be conflated. It also allows the same trained model to be viewed as an unconditional generator or an incomplete travel diary infilling model.

\section{Canonical Diary Encoding}
\label{sec:encoding}
In this section, we describe how we link household, person, and travel surveys in a way that is digestible by masked generative models such as autoregressive transformers and discrete diffusion models. We define the canonical encoding that maps one raw travel diary to a single fixed-length token sequence. The design goals are to keep the household-day whole, to make its variable structure explicit through dedicated tokens rather than implicit in the sequence length, and to distinguish the different kinds of absence that a padded, partially observed survey record contains.
 
\begin{figure*}[t]
  \centering
  \includegraphics[width=\textwidth]{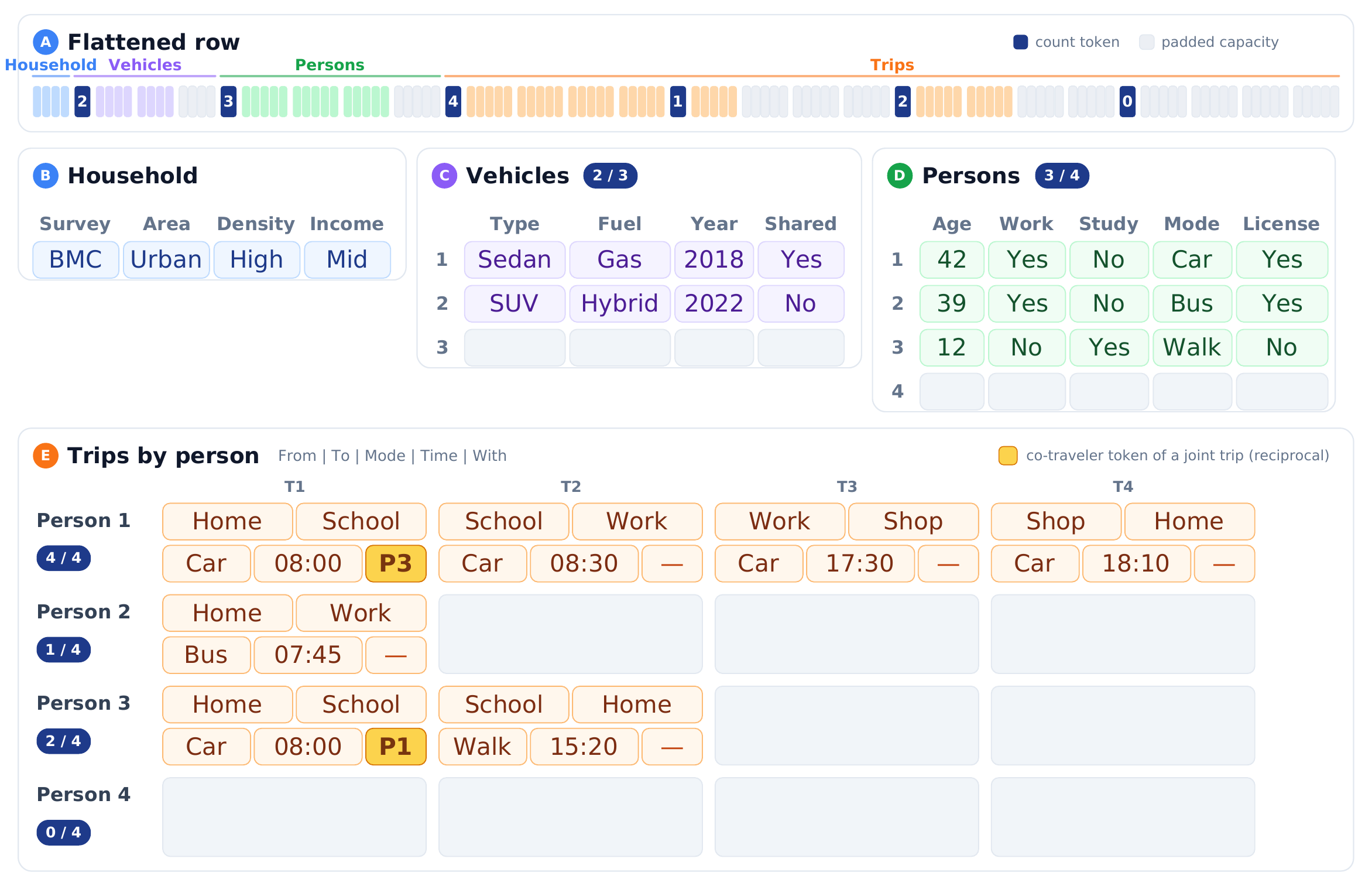}
  \caption{The top ribbon shows one encoded sample in flattened canonical order. The lower panels unfold the same sample into household, vehicle, person, and person-wise trip blocks. The example is illustrative: it uses reduced caps (Vmax = 3, Pmax = 4, Tmax = 4) and simplified attribute labels; the attributes used in the experiments are listed in Table \ref{tab:features}.}
  \label{fig:canonical-sample-clean}
\end{figure*}
 
\subsection{The raw diary object}
\label{sec:raw}
 
The raw object for one household-day is
\begin{equation}\label{eq:raw-data}
     x_\text{raw} = \big(h, V_{1:n_V}, P_{1:n_P}, (T_i)_{i=1}^{n_P}\big),
     \qquad
     T_i = (t_{i,1}, \dots, t_{i,\ell_i}),
\end{equation}
where $h$ is the household block, $V_{1:n_V} = (v_1, \dots, v_{n_V})$ is the set of household vehicles, $P_{1:n_P} = (p_1, \dots, p_{n_P})$ is the set of persons in the household, and $T_i$ is the ordered trip chain of person $i$. Each $v_j$, $p_i$, and $t_{i,k}$ is itself a block of attributes rather than a single scalar token. The challenge is that $n_V$, $n_P$, and the per-person trip counts $\ell_i$ vary across households.
 
\subsection{Canonical padded form}
\label{sec:canonical}
 
We define a canonical padded encoding $x = E(x_\text{raw})$ with the form
\begin{equation}\label{eq:data}
\begin{split}
x = \big[ h,&c_V, V_1, \dots, V_{V_{\max}},c_P, P_1, \dots, P_{P_{\max}}, \\
&c_{T,1}, T_{1,1}, \dots, T_{1,T_{\max}},\dots,c_{T,P_{\max}}, T_{P_{\max},1}, \dots, T_{P_{\max},T_{\max}} \big],
\end{split}
\end{equation}
where $c_V=n_V$, $c_P=n_P$, and $c_{T,i}=\ell_i$ are explicit count tokens, and $V_{\max}$, $P_{\max}$, and $T_{\max}$ are capacity caps on the vehicle roster, the person roster, and each person's trip chain. All slots beyond the realized counts are filled with a dedicated padding symbol, so the encoding pads rather than crops: any household within the caps is represented exactly, and one fixed-length structured sample preserves the block boundaries of the original hierarchy. Figure~\ref{fig:canonical-sample-clean} unfolds one encoded sample. With $|h|$, $|v|$, $|p|$, and $|t|$ attributes per block, every encoded diary has the same length
\begin{equation}\label{eq:length}
L = |h| + 1 + V_{\max}|v| + 1 + P_{\max}|p| + P_{\max}\big(1 + T_{\max}|t|\big).
\end{equation}
 
Let $\mathcal{C}$ denote the $2 + P_{\max}$ count positions. The count tokens and the padding pattern are deterministically coupled: given the counts, the set $\mathcal{A}$ of realized (active) slots is fixed, namely the first $c_V$ vehicle blocks, the first $c_P$ person blocks, and the first $c_{T,i}$ trip blocks of each realized person $i$, and every position outside $\mathcal{A} \cup \mathcal{C}$ must equal $\pad$. We call an encoded sequence that satisfies this \emph{count--padding coupling} structurally consistent. The coupling is the invariant that the count-first sampler of Section~\ref{sec:count-first} guarantees by construction.
 
\subsection{Special-token semantics}
\label{sec:special-tokens}
 
Inactive slots are not confused with genuinely unobserved or inapplicable fields. The encoding reserves four special symbols with distinct semantics: $\pad$ marks a position outside the realized object, $\mask$ marks a position intentionally hidden during denoising, $\missing$ marks a field not observed in the survey, and $\na$ marks a field that is structurally inapplicable. $\missing$ and $\na$ are ordinary vocabulary entries at the positions where they can occur, so the model learns the survey's missingness patterns; $\mask$ is kept outside the data vocabulary and exists only as the corruption symbol of the diffusion process (Section~\ref{sec:mdm}). Whether the \emph{sampler} may emit $\missing$ is a policy switch, made explicit here because it changes what the synthetic data means: for population synthesis we suppress $\missing$ at sampling time through the legality mask of Section~\ref{sec:legality}, so that generated diaries are complete.
 
\subsection{Position-dependent vocabularies}
\label{sec:vocab}
 
A token that is meaningful at one position is meaningless at another: household-income values are not departure-time bins. Each position $\ell$ therefore carries its own admissible set $\mathcal{V}_\ell$. Field positions admit their field's value vocabulary plus $\missing$ or $\na$ where applicable, count positions admit exactly the integers $0, \dots, \text{cap}$ of the roster they govern, and $\pad$ is admissible at every content position because whether a slot is active depends on the counts. The global vocabulary $\mathcal{V}$ is the union of all per-field vocabularies, the count values, and the special tokens. Section~\ref{sec:legality} enforces admissibility during training and sampling.
 
One trip attribute deserves separate discussion because it encodes joint travel. We introduce a novel \emph{co-traveler} token: a bitmask over the $P_{\max}$ person slots marking which household members accompanied the traveler, with the traveler's own bit excluded by construction. The trip's position in the encoding identifies the traveler, so a trip inside person block $i$ admits only bitmasks whose bit $i$ is zero, and a traveler can never be recorded as accompanying themselves. This token is what makes within-household joint travel a measurable property of generated diaries (Sections \ref{sec:res-member-evidence} and \ref{sec:res-feas}).

\section{Methodology: DiaryBridge}
\label{sec:method}
This section introduces DiaryBridge, a masked diffusion model over the tokenized travel diaries.
 
\subsection{Masked discrete diffusion over structured encodings}
\label{sec:mdm}
 
Let $x = (x^{(1)}, \dots, x^{(L)})$ denote an encoded diary as defined in Section~\ref{sec:encoding}. The forward process of masked (absorbing-state) diffusion replaces tokens independently with the $\mask$ symbol: given a decreasing schedule $\alpha_t$ with $\alpha_0 \approx 1$ and $\alpha_1 \approx 0$,
\begin{equation}
q(z_t \mid x) = \prod_{\ell=1}^{L} q\big(z_t^{(\ell)} \mid x^{(\ell)}\big), \qquad z_t^{(\ell)} = 
\begin{cases}
    x^{(\ell)} & \text{with probability } \alpha_t,\\
    \mask & \text{with probability } 1-\alpha_t.
\end{cases}
\end{equation}
A bidirectional transformer $p_{\vtheta}(\cdot \mid z_t)$ predicts the clean token at every masked position. Following the simplified continuous-time formulation of masked diffusion \citep{shi2024md4,sahoo2024mdlm}, we train with
\begin{equation}\label{eq:nelbo}
\mathcal{L}(\vtheta) = \sE_{x \sim \pdata} \sE_{t \sim \mathcal{U}(0,1)} \sE_{z_t \sim q(\cdot \mid x)} \bigg[ \frac{\alpha_t'}{1-\alpha_t} \sum_{\ell: z_t^{(\ell)} = \mask} \log p_{\vtheta}\big(x^{(\ell)} \mid z_t\big) \bigg],
\end{equation}
which is the (rescaled) negative evidence lower bound (NELBO); since $\alpha_t' \le 0$, the objective is a nonnegative weighted cross-entropy over masked positions. 

Sampling proceeds by ancestral reverse steps on a time grid $1 = t_K > \dots > t_0 = 0$: at each step from $t$ to $s < t$, every masked position is independently unmasked with probability $(\alpha_s - \alpha_t)/(1-\alpha_t)$, drawing the revealed token from $p_{\vtheta}(\cdot \mid z_t)$ \citep{shi2024md4}. Training masks positions uniformly at random across the entire encoding, including count tokens and $\pad$ positions, so the model learns the joint law of structure and content.
 
Conditioning on observed tokens requires no architectural change: for any observed set $X_{\mathrm{obs}}$, the sampler initializes those positions with their observed values instead of $\mask$, and the reverse process, which only ever modifies masked positions, keeps them clamped throughout. Unconditional synthesis and imputation of arbitrary fields are therefore the same procedure with different initial masks.
 
\subsection{Legality masking}
\label{sec:legality}
 
Section~\ref{sec:vocab} assigns each position $\ell$ an admissible set $\mathcal{V}_\ell \subset \mathcal{V}$. We enforce admissibility with a fixed legality mask $M \in \{0,1\}^{L \times |\mathcal{V}|}$, where $M_{\ell k} = 1$ if and only if token $k$ belongs to $\mathcal{V}_\ell$. The mask is enforced by restricting the categorical distribution at every position,
\begin{equation}\label{eq:legality}
p_{\vtheta}\big(x^{(\ell)} = k \mid z_t\big) = \frac{M_{\ell k} \exp f_{\vtheta}^{(\ell,k)}(z_t)}{\sum_{k'} M_{\ell k'} \exp f_{\vtheta}^{(\ell,k')}(z_t)},
\end{equation}
in the \emph{training loss} of Eq.~\eqref{eq:nelbo} as well as in the sampler. Beyond value spaces, the mask carries the structural restrictions of Section~\ref{sec:vocab}, such as the exclusion of self-referential co-traveler bitmasks, and it implements sampling-time policy: suppressing $\missing$ for population synthesis (Section~\ref{sec:special-tokens}) amounts to zeroing the corresponding entries of $M$ at sampling time.
 
\subsection{Count-first sampling}
\label{sec:count-first}
\begin{figure*}[t]
  \centering
  \includegraphics[width=\textwidth]{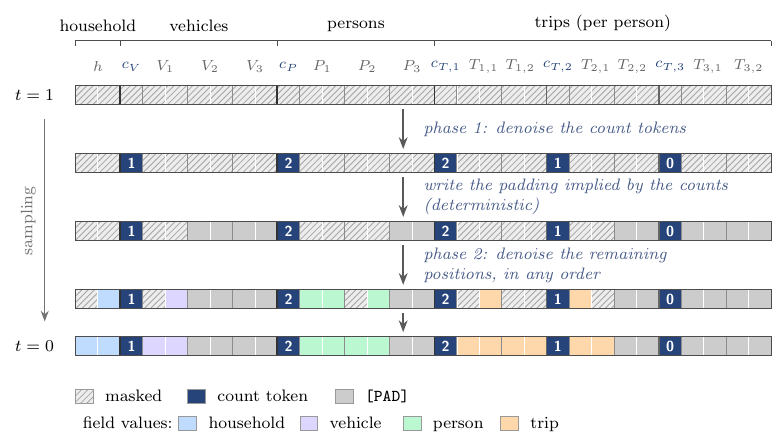}
  \caption{Simplified illustration of the two-phase count-first discrete diffusion sampler.}
  \label{fig:count-first}
\end{figure*}
Section~\ref{sec:canonical} defines the count--padding coupling: the counts determine the active-slot set $\mathcal{A}$, and every position outside $\mathcal{A} \cup \mathcal{C}$ must equal $\pad$. Sampling all positions jointly cannot guarantee this coupling. Rather than adding an auxiliary structure model or repairing samples after the fact, we exploit the fact that masked diffusion admits arbitrary unmasking orders: we simply schedule the counts first.
 
\begin{algorithm}[t]
\caption{Count-first sampling with optional conditioning}
\label{alg:count-first}
\begin{algorithmic}[1]
\REQUIRE denoiser $p_{\vtheta}$, legality mask $M$, count positions $\mathcal{C}$, observed tokens $x_{\mathrm{obs}}$ with indicator $m$ (possibly empty), step budgets $K_1, K_2$
\STATE $z^{(\ell)} \gets x_{\mathrm{obs}}^{(\ell)}$ if $m^{(\ell)} = 1$, else $z^{(\ell)} \gets \mask$, for all $\ell$
\FOR{$k = 1$ \TO $K_1$}
  \STATE ancestral reverse step in which the unmasking probability is set to zero at every position outside $\mathcal{C}$
\ENDFOR
\STATE read the counts from $z$ and form the active-slot set $\mathcal{A}$
\STATE $c_{T,i} \gets 0$ for every $i > c_P$
\STATE $z^{(\ell)} \gets \pad$ for every unobserved position $\ell \notin \mathcal{A} \cup \mathcal{C}$
\FOR{$k = 1$ \TO $K_2$}
  \STATE standard ancestral reverse step on the remaining masked positions
\ENDFOR
\RETURN $z$
\end{algorithmic}
\end{algorithm}
 
Algorithm~\ref{alg:count-first} runs in two phases. Phase one restricts unmasking to the $2 + P_{\max}$ count positions, so the counts are drawn from their (model) joint distribution given whatever is observed. The padding pattern implied by the sampled counts is then written deterministically, and $\pad$ is never itself a sampling outcome. Phase two denoises only the realized slots. By construction, every sample satisfies the count--padding coupling exactly, so the structural-violation rate of DiaryBridge under count-first sampling is zero. Figure \ref{fig:count-first} illustrates the two-stage count-first sampling. The interesting empirical question, which our evaluation tests, is whether the two-phase schedule costs anything in distributional fit relative to joint sampling, the single-phase sampler that treats $\pad$ as an ordinary token.

\section{Data and Experimental Setup}
\label{sec:experiments}

We evaluate linked-table synthesis, household travel dependencies, and travel generation. The data and comparison methods are shared across the study; each experiment's design, measures, and results are presented together in Section~\ref{sec:results}.

\subsection{Data}
\label{sec:data}

\subsubsection{Survey and linked diary object}
 
We use the household, person, vehicle, and trip tables of the 2017/2018 Regional Travel Survey (RTS), a household travel survey of the metropolitan Washington--Baltimore region conducted by the \citet{TPB2021RTS} in cooperation with the Baltimore Metropolitan Council (BMC). This paper uses the BMC-region extract of the public files. We link the four tables on household and person identifiers into one diary object $x_\text{raw}$ per household as in Eq.~\eqref{eq:raw-data}.
 
Table~\ref{tab:data-stats} summarizes the linked data. Persons with no reported trips (2{,}909 persons, 16.3\%) are included in the trips-per-person statistics with a count of zero, and households with no vehicles (655 households, 7.9\%) are included in the vehicles-per-household statistics with a count of zero; both are legitimate outcomes that the model must represent, encoded as a count token of $0$ with all corresponding slots set to $\pad$. 

\begin{table}[!htbp]
\centering
\caption{Linked BMC extract of the 2017/2018 RTS. Records, means, and percentiles describe all 8{,}313 household-days; the last column gives the records retained after the capacity caps of Section~\ref{sec:caps}.}
\label{tab:data-stats}
\begin{tabular}{lrrrrr}
\toprule
 & Records & Mean & 99th pct. & Max & After caps \\
\midrule
Households & 8{,}313 & --- & --- & --- & 8{,}201 \\
Persons per household & 17{,}858 & 2.15 & 6 & 10 & 17{,}552 \\
Vehicles per household & 15{,}311 & 1.84 & 5 & 12 & 14{,}913 \\
Trips per person & 59{,}913 & 3.35 & 11 & 41 & 57{,}759 \\
\bottomrule
\end{tabular}
\end{table}
 
\subsubsection{Capacity caps}
\label{sec:caps}
 
Guided by the 99th percentiles in Table~\ref{tab:data-stats}, we set $P_{\max}=8$, $V_{\max}=6$, and $T_{\max}=12$. Only 2 households exceed $P_{\max}$, 28 exceed $V_{\max}$, and 87 person-day trip chains (located in 82 households) exceed $T_{\max}$. We drop any household that exceeds a cap in any dimension, removing 112 households (1.35\%) and leaving 8{,}201 households with 17{,}552 persons, 14{,}913 vehicles, and 57{,}759 trips.
 
\subsubsection{Feature selection, discretization, and missing values}
\label{sec:features}
 
Each block contributes a fixed set of attributes, listed in Table~\ref{tab:features}. With $|h|=4$ household, $|v|=4$ vehicle, $|p|=5$ person, and $|t|=5$ trip attributes, Eq.~\eqref{eq:length} gives the encoded length $L = 4 + 1 + 24 + 1 + 40 + 8(1+60) = 558$. The resulting global vocabulary has 299 entries: the field values, the count values, the 123 distinct co-traveler patterns observed in the extract (including traveling alone), and the special tokens, with $\mask$ kept outside it (Section~\ref{sec:special-tokens}).
 
Numeric attributes are discretized. We use 30-minute departure-time bins (48 bins per day), age bins of $0$--$4$, $5$--$15$, $16$--$17$, $18$--$24$, and ten-year bins thereafter up to $75{+}$, and vehicle model-year bins of five years with a single pre-1990 bucket. We empirically analyzed the bin-width sensitivity to compare 15-, 30-, and 60-minute discretizations: departure-time total variation distance on the validation split is $0.062$ at 30 minutes versus $0.102$ at 15 and $0.121$ at 60, hence the 30-minute default.
 
Reserved survey codes for refused, don't-know, and not-ascertained responses, identified from the public-file data dictionary \citep{TPB2021RTS}, are mapped to $\missing$; structurally inapplicable fields are mapped to $\na$ where the dictionary distinguishes them.
 
\begin{table}[!htbp]
\centering
\caption{Feature selection.}
\label{tab:features}
\footnotesize
\begin{tabular}{l p{0.60\textwidth} c}
\toprule
Block & Attributes (type) & Slots \\
\midrule
Household & area type, home type, home ownership, household income (cat.) & $1 \times 4$ \\
Vehicle & body type, fuel type, model year (binned), toll transponder (cat.) & $V_{\max} \times 4$ \\
Person & age (binned), gender, employment status, license, disability (cat.) & $P_{\max} \times 5$ \\
Trip & origin activity, destination activity, mode (cat.), departure time (binned), co-travelers (bitmask) & $P_{\max} T_{\max} \times 5$ \\
Counts & $c_V$, $c_P$, $c_{T,1},\dots,c_{T,P_{\max}}$ & $2 + P_{\max}$ \\
\bottomrule
\end{tabular}
\end{table}
 
\subsection{Training and comparison methods}
\label{sec:training}
\label{sec:baselines}

The household split contains 6{,}561 training, 820 validation, and 820 test records. DiaryBridge is selected by validation NELBO (Eq. \ref{eq:nelbo}) from a grid of transformer depths, widths, and dropout rates. The selected model has four layers, 384-dimensional features, six attention heads, and dropout 0.1 (12.3M parameters), and at a 40k-step budget early stopping fires near 25k steps for every leading configuration, so the reported models are converged rather than budget-limited. Diffusion sampling uses temperature one, 64 count steps, and 512 content steps.

Empirical resampling and IPF/IPU with training controls provide diary-reuse references \citep{Beckman1996Baseline,Ye2009IPU}. They preserve relationships within the training diaries but cannot introduce household patterns absent from training. They establish what survey reuse achieves without learning a new generator.

Bayesian networks (BN) are an established approach to population and activity-chain synthesis \citep{Sun2015BN,Joubert2020ActivityBN}. Their explicit dependencies and compact local distributions make BN a strong reference for demographic fit and sequential travel generation. Our hierarchical BN baseline extends \citet{Sun2015BN}: it generates household attributes and counts, persons and vehicles conditional on the household, and each person's trips through a first-order chain. Structure learning uses hill climbing, at most three parents per node, and Laplace smoothing. Origin--destination activity continuity is enforced; members are conditionally independent given household information. Adding parents allows richer dependencies, but tabular conditional distributions grow multiplicatively with parent cardinalities and require more parameters to be estimated \citep{Sun2015BN}.

The VAE \citep{Borysov2019MicroAgents} and CTGAN \citep{Xu2019CTGAN} provide neural alternatives for population and tabular synthesis. Applied to the flattened token encoding, they test whether generic neural generators can reproduce linked travel surveys.

Our capacity-matched autoregressive transformer (AR) provides the closest comparison, sharing the canonical encoding, legality masking, count-triggered padding, and legal co-traveler masks. AR can learn cross-member dependencies, but its native completion procedure reads observed tokens in canonical left-to-right order and uses only earlier tokens at hidden positions. Comparing it with DiaryBridge tests left-to-right against any-order unmasking. 

DiaryBridge's \emph{count-first} sampler uses Algorithm~\ref{alg:count-first}. DiaryBridge's \emph{joint} sampler instead denoises counts and content together, treating $\pad$ as an ordinary token. Count-first and AR enforce count--padding consistency; joint has no such guarantee. Evaluating both samplers on the same checkpoints assesses the fidelity--validity trade-off of the count-first sampler.

For conditional generation, the \emph{diary donor} tests whether transferring travel from similar training households is sufficient. It samples uniformly from the ten closest training households with the same roster size. Distance combines household categorical mismatches, vehicle-count differences divided by three, and mean per-slot demographic mismatches. Observed trip-count differences are divided by the trip cap and weighted by two; peer-trip mismatch contributes only when peer diaries are observed. Ties are randomized, and hidden target attributes never enter matching. This reference reuses training diaries, so its aggregate accuracy does not establish novel-record generation or privacy protection.

Results with uncertainty report means and sample standard deviations across five model/sampling seeds. Table rankings compare learned generators, excluding real-data and diary-reuse references.

\section{Experiments and Results}
\label{sec:results}
The experiments address two questions. First, does modeling the household-day as a single linked object produce more realistic travel diaries, especially when only part of a diary is observed? Second, how does DiaryBridge compare with established synthesis approaches, from survey reuse and Bayesian networks to deep generative models and an autoregressive transformer sharing the same encoding? No single statistic captures what makes a set of synthetic diaries useful, so each subsection evaluates a complementary property: the fit of joint distributions across the household, person, and trip tables (Section~\ref{sec:res-fit}); whether modes and purposes reflect household resources and employment status (Section~\ref{sec:resources}); whether trips shared between household members are reproduced consistently (Section~\ref{sec:res-member-evidence}); whether generated travel reproduces trip volumes and their distribution across modes, purposes, and departure periods for a given population (Section~\ref{sec:res-demand}); whether synthetic records support predictive models trained on them (Section~\ref{sec:res-downstream}); and whether generated records are structurally valid and cover plausible household patterns absent from training (Section~\ref{sec:res-feas}). These experiments target the uses of synthetic diaries most relevant to travel demand analysis rather than every possible validation.

\subsection{General synthesis fidelity}
\label{sec:res-fit}

Matching household and trip distributions separately can conceal errors in the relationships between travel and traveler characteristics. We therefore compare joint distributions across the linked tables. Each method generates 8{,}201 unconditional households, which are compared with the real test split.

The household–person comparison pairs a household attribute or vehicle count with a person attribute, with each person contributing one value pair. The person–trip comparison pairs a traveler attribute with a trip attribute, with each trip contributing one value pair. Household--person--trip triples contain exactly one attribute from each table, such as income, employment status, and mode; vehicle count is included only in the household--person comparison. Finally, member pairs compare the same attribute across two distinct household members, such as their age groups, and average over the five person attributes. These value pairs are unordered, so this last comparison measures household composition rather than member-slot order.

For each attribute set $S$, we compare real and generated relative frequencies, $\pi_{S,c}$ and $\hat\pi_{S,c}$, over its categorical cells $C_S$. The support includes $\missing$ and, where allowed, $\na$. Standardized root mean squared error is
\begin{equation}
\label{eq:srmse-eval}
\mathrm{SRMSE}(S)
=\sqrt{|C_S|\sum_{c\in C_S}\big(\hat\pi_{S,c}-\pi_{S,c}\big)^2}.
\end{equation}
This is RMSE divided by the mean real cell frequency \citep{Sun2015BN,Borysov2019MicroAgents,Kim2023FeasibleDiverse}; zero means identical frequencies. Pairs ($k=2$) and triples ($k=3$) remain separate; their averages summarize the specified relationships, not the full diary distribution.

Each entry of Table~\ref{tab:srmse-cross} is therefore an average SRMSE over the attribute combinations of one family. Values are comparable within a column but not across columns, because families differ in their numbers of cells and triples are sparser than pairs. The Resample and IPF/IPU rows, which reuse real training households, indicate the error attributable to sampling noise alone: with 820 test households, even real diaries score between .14 and .74. A generator should therefore be judged by its distance to these rows rather than to zero.

BN has the lowest household--person error among learned generators, while joint DiaryBridge leads on traveler--trip, household--person--trip, and member-pair relationships (Table~\ref{tab:srmse-cross}). AR is second on member pairs, supporting the shared household-day encoding beyond diffusion alone. Joint outperforms count-first in every family, although distributional fit alone does not establish which sampler is more useful for completion or produces valid records.

\begin{table}[!htbp]
\centering
\caption{Cross-panel SRMSE at the indicated feature orders (lower is better; mean $\pm$ std). Bold and underlined entries mark the best and second-best learned generators, respectively. The number of attributes is $k$.}
\label{tab:srmse-cross}
\footnotesize
\setlength{\tabcolsep}{4pt}
\begin{tabular}{lcccc}
\toprule
Method & Household & Person & Household-- & Member \\
 & $\times$ person & $\times$ trip & person--trip & pairs \\
 & $k=2$ & $k=2$ & $k=3$ & $k=2$ \\
\midrule
Resample & $.138\pm.005$ & $.359\pm.016$ & $.744\pm.027$ & $.165\pm.009$ \\
IPF/IPU (train controls) & $.135\pm.008$ & $.353\pm.008$ & $.743\pm.026$ & $.163\pm.009$ \\
\midrule
BN & $\mathbf{.229\pm.015}$ & $\underline{.921\pm.009}$ & $\underline{1.646\pm.026}$ & $.583\pm.091$ \\
VAE & $.481\pm.065$ & $1.168\pm.095$ & $2.154\pm.240$ & $.883\pm.051$ \\
CTGAN & $1.330\pm.061$ & $4.369\pm.082$ & $6.678\pm.201$ & $1.945\pm.114$ \\
AR & $.437\pm.031$ & $2.098\pm.270$ & $2.963\pm.331$ & $\underline{.374\pm.048}$ \\
DiaryBridge count-first & $.554\pm.097$ & $.988\pm.098$ & $1.926\pm.265$ & $.413\pm.140$ \\
DiaryBridge joint & $\underline{.343\pm.061}$ & $\mathbf{.621\pm.081}$ & $\mathbf{1.326\pm.170}$ & $\mathbf{.271\pm.043}$ \\
\bottomrule
\end{tabular}
\end{table}

\subsection{Resources and traveler roles}
\label{sec:resources}
This section asks the following question: when the population and number of trips are fixed, does the model assign modes and purposes that reflect household resources and employment status? Matching overall mode and purpose shares does not ensure that travel is assigned appropriately across households and travelers. We therefore test whether generated mode shares reflect household vehicle resources and whether trip purposes reflect employment status. To separate these relationships from errors in travel volume, we hold all household, vehicle, and person attributes fixed, together with each person’s observed trip count. Each method generates only the trip attributes for the same 820 test households, comprising 1,771 persons and 6,055 trips.

We compare mode distributions within three vehicle-sufficiency groups: no vehicles, a positive vehicle count below the number of recorded licensed members, and sufficient vehicles. Purpose distributions are compared within worker, nonworker, employment-inapplicable, and employment-missing groups. For context $G$ and outcome $Y$, trip-weighted conditional total variation distance measures how much the generated distributions differ from the real distributions \emph{within} these groups:
\begin{equation}
\label{eq:conditional-tvd}
\mathrm{TVD}(Y\mid G)
=\sum_g w_g\frac{1}{2}\sum_y
\big|\hat p(y\mid g)-p(y\mid g)\big|,
\qquad w_g=\frac{n_g}{N},
\end{equation}
Here, $n_g$ is the group's real trip count and $N=\sum_g n_g=6{,}055$. The score ranges from zero for identical distributions to one for disjoint distributions. It measures the probability mass requiring reassignment, averaged using real trip shares.

Because small groups receive less weight in this average, two targeted rates complement it. The first is the percentage of the 68 zero-vehicle households receiving a driver trip. The second is the percentage of generated work-destination trips attributed to nonworkers. Their denominators are households and work trips, respectively. Both are empirical associations to reproduce, not hard structural prohibitions.

Count-first has the lowest conditional TVD among learned generators for both mode given resources ($0.040$) and purpose given employment ($0.044$; Table~\ref{tab:transport-context}). Its resource-conditioned mode fit is close to the donor's $0.035$ and better than BN's $0.095$. Thus, its advantage extends beyond matching pooled travel shares to assigning modes and purposes according to observed household context.

\begin{table}[!htbp]
\centering
\caption{Travel conditional on resources and employment (mean $\pm$ SD). Lower TVD and empirical rates closer to the real reference are better. Bold/underline: best/second learned generator.}
\label{tab:transport-context}
\footnotesize
\setlength{\tabcolsep}{3.5pt}
\begin{tabular}{lcccc}
\toprule
 & \multicolumn{2}{c}{Conditional TVD} & \multicolumn{2}{c}{Empirical rate (\%)} \\
\cmidrule(lr){2-3}\cmidrule(lr){4-5}
Method & Mode given & Purpose given & Zero-vehicle & Work trips by \\
 & resources & employment & households with & nonworkers \\
 & & & driver trips & \\
\midrule
Real & --- & --- & 0.0 & 2.8 \\
Diary donor & $.035\pm.009$ & $.050\pm.006$ & $14.1\pm3.0$ & $19.6\pm1.1$ \\
\midrule
BN & $\underline{.095\pm.005}$ & $\underline{.060\pm.006}$ & $69.7\pm2.2$ & $9.8\pm0.8$ \\
AR & $.119\pm.004$ & $.080\pm.010$ & $4.7\pm4.5$ & $\mathbf{4.7\pm2.2}$ \\
DiaryBridge count-first & $\mathbf{.040\pm.009}$ & $\mathbf{.044\pm.005}$ & $\underline{3.2\pm4.1}$ & $10.5\pm1.6$ \\
DiaryBridge joint & $.155\pm.180$ & $.155\pm.177$ & $\mathbf{2.6\pm1.2}$ & $\underline{9.3\pm3.5}$ \\
\bottomrule
\end{tabular}
\end{table}

The targeted rates qualify this overall result. Count-first generates driving in $3.2\%$ of zero-vehicle households, versus $69.7\%$ for BN and none in the real sample. AR is closest to the real nonworker share of work trips: $4.7\%$, compared with $2.8\%$ real and $10.5\%$ count-first. These rankings answer different questions: purpose TVD compares whole purpose distributions within employment groups, whereas the targeted rate asks who makes work trips.

\subsection{Missing-member recovery}
\label{sec:res-member-evidence}

We next test whether observing other household members’ travel diaries improves recovery of shared trips in a missing member’s diary. Household characteristics and trip counts provide context, but peer diaries can provide evidence of specific joint travel. For example, a child’s passenger trip naming a parent provides evidence of a compatible trip in the parent’s diary. We therefore compare completion with and without observed peer diaries, evaluating whether the generated target diary reproduces these shared-trip relationships.

The cohort contains all 534 test households with at least two members and one traveler. We select one traveling member uniformly within each household as the target. In the without-peer-diaries condition, all members’ active trip attributes are hidden. In the with-peer-diaries condition, the other members’ trip diaries remain observed, and only the target’s trip attributes are hidden. Both conditions retain all household, vehicle, and person attributes, all trip counts, and the inactive padding positions. This comparison isolates the benefit of making peer information available to each existing completion procedure, without retraining the models.

We measure recovery through reciprocal reports of shared travel. The real peer diaries contain 892 references to the selected targets across 228 households: each reference is a peer trip naming the target as a companion. A reference is recovered when the generated target diary contains a trip that names the peer in return, has a compatible mode, and departs in the same or an adjacent 30-minute bin. Driver and passenger modes also need to be compatible. Recovery is the percentage of the 892 real references with such a counterpart. This fixed denominator prevents a method from improving its score simply by generating fewer companion claims.

\begin{figure}[!htbp]
\centering
\includegraphics[width=\linewidth]{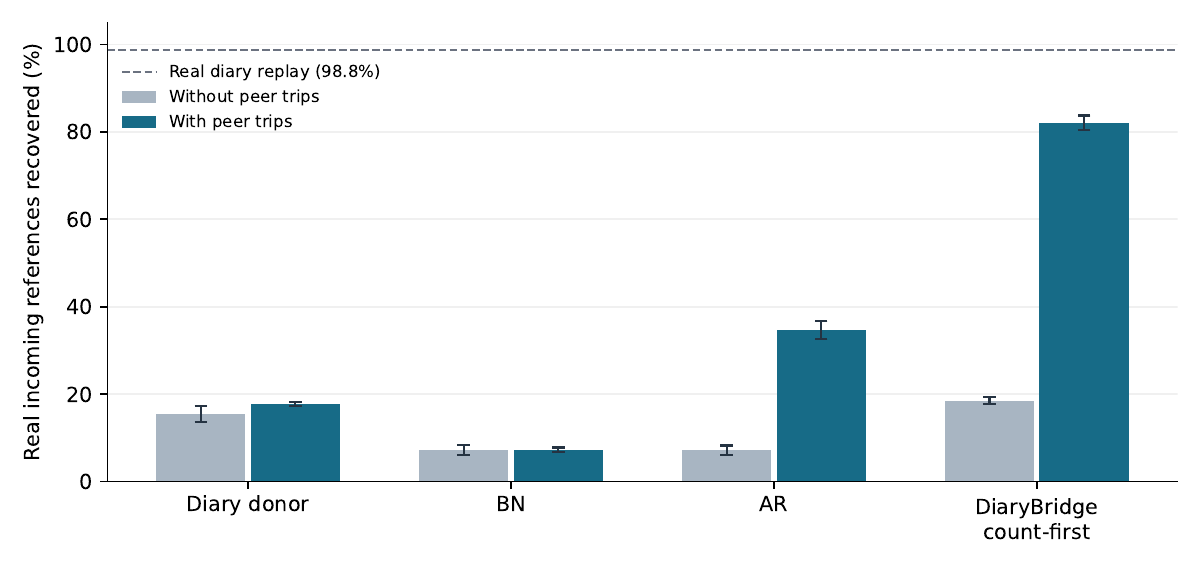}
\caption{Missing-member recovery with and without observed peer diaries. Recovery is the percentage of 892 real peer references with a compatible generated target trip. Bars show five-run means and error bars sample SD; attributes and trip counts are observed in both conditions.}
\label{fig:member-evidence}
\end{figure}

Applying the recovery rule to the real target diaries recovers 881 of the 892 references, or 98.8\%. Revealing peer diaries increases DiaryBridge recovery from 18.5\% to 82.1\% (Figure \ref{fig:member-evidence}). AR also benefits, increasing from 7.2\% to 34.6\%. Donor recovery changes only from 15.5\% to 17.8\%. BN remains near 7.3\%. Recovery alone could reward excessive companion claims because additional claims can increase the number of matched real references, while unsupported claims do not reduce the score. We therefore also report precision, the percentage of generated target-to-peer references supported by a compatible trip in the real peer diary. With peer diaries observed, precision is also higher for DiaryBridge than for AR: 80.4\% versus 52.7\%. DiaryBridge therefore recovers more real references while generating a higher proportion of supported companion claims.

\subsection{Travel-demand application}
\label{sec:res-demand}
We test whether each method can assign travel to a given population while reproducing both the number of trips and their distribution across modes, purposes, and departure periods. Matching travel shares alone is insufficient: a method could reproduce those shares while generating too few trips. We therefore retain the household, vehicle, and person records of the 820 test households, comprising 1,771 persons, and generate all trip counts and trip attributes. 

We first assess travel volume through the mean number of trips per person. The generated trip total is divided by the fixed population of 1,771 persons, including those assigned no trips. The real reference is 3.419 trips per person. This measure indicates whether a method generates enough travel overall.

Then, to assess volume and composition together, we compare real and generated trip counts across travel groups defined using mode, destination purpose, and departure period. The grouping uses seven mode categories, six purpose categories, and five departure periods. Let \(N_c\) and \(\widehat N_c\) denote the real and generated trip counts in travel group \(c\). We calculate
\[
E_{\mathrm{demand}} = 100 \frac{\sum_c \big|\widehat N_c-N_c\big|}{N},
\]
where $N=\sum_cN_c=6{,}055$ is the real trip total for the fixed population. The numerator adds the absolute shortfalls and excesses across groups, so errors in different groups do not cancel. Moving a trip between groups contributes both a shortfall in one group and an excess in another. Note that lower scores are better, and that the score can exceed 100\%. This comparison assesses aggregate demand by travel characteristics. All methods preserve the observed population; joint requires count–padding adjustments in 10.1\% of households before scoring, versus none for the other methods.

Table \ref{tab:transport-demand} shows that BN comes closest to reproducing total travel volume among learned generators. The combined demand score gives a different ranking. Joint DiaryBridge has the lowest demand error among learned generators, at 28.67\%, compared with 39.95\% for BN. This distinction matters: reproducing the total number of trips does not ensure that those trips are distributed appropriately across the evaluated travel groups. DiaryBridge joint achieves a better combined volume-and-composition score.

\begin{table}[!htbp]
\centering
\caption{Travel generated for a fixed population of 820 households (mean $\pm$ SD). Trips/person is ranked by closeness to real data; lower demand error is better. Bold/underline: best/second learned generator.}
\label{tab:transport-demand}
\small
\begin{tabular}{lcc}
\toprule
Method & Trips/person & Demand error (\%) \\
\midrule
Real & 3.419 & --- \\
Diary donor & $3.407\pm0.061$ & $16.55\pm0.68$ \\
\midrule
BN & $\mathbf{3.278\pm0.062}$ & $39.95\pm1.62$ \\
AR & $1.700\pm0.126$ & $52.92\pm3.87$ \\
DiaryBridge count-first & $2.306\pm0.340$ & $\underline{37.38\pm8.57}$ \\
DiaryBridge joint & $\underline{2.752\pm0.246}$ & $\mathbf{28.67\pm4.70}$ \\
\bottomrule
\end{tabular}
\end{table}

\subsection{Downstream predictive utility}
\label{sec:res-downstream}
In this section, we test whether synthetic records preserve information useful for prediction on real data. We train identical gradient-boosted classifiers for two tasks: predicting trip mode from household, traveler, time, and activity attributes, and predicting vehicle-count class from household attributes and household size. The classifiers are trained on synthetic records and evaluated on held-out real records

For each task, we compare the classifier’s macro-F1 score with the score obtained by training on real data. Macro-F1 gives each outcome class equal weight. A ratio of one matches the real-training reference, while a ratio below one indicates lower predictive performance. 

DiaryBridge has the highest trip-mode F1 ratio among learned generators (Table \ref{tab:predictive}). DiaryBridge joint performs best on vehicle-count prediction followed by BN. These results indicate task-specific predictive utility rather than a uniform advantage for one generator. 

\begin{table}[!htbp]
\centering
\caption{Train-on-synthetic, test-on-real macro-F1 relative to training on real data (higher is better). Bold/underline: best/second learned generator.}
\label{tab:predictive}
\small
\begin{tabular}{lcc}
\toprule
Method & Trip mode & Vehicle count \\
\midrule
Resample & .945 & .999 \\
IPF/IPU (train controls) & .976 & 1.001 \\
\midrule
BN & .613 & $\underline{1.020}$ \\
VAE & .737 & .690 \\
CTGAN & .522 & .431 \\
AR & $.833$ & .971 \\
DiaryBridge count-first & $\underline{.963}$ & $.992$ \\
DiaryBridge joint & $\mathbf{1.020}$ & $\mathbf{1.051}$ \\
\bottomrule
\end{tabular}
\end{table}

\subsection{Structural validity and support coverage}
\label{sec:res-feas}

\begin{table}[!htbp]
\centering
\caption{Raw encoded errors, activity-chaining errors, and sampling-zero coverage (\%, mean $\pm$ SD). Lower errors and higher coverage are better. Bold/underline: best/second learned generator.}
\label{tab:feasibility}
\footnotesize
\setlength{\tabcolsep}{4pt}
\begin{tabular}{lcccc}
\toprule
 & \multicolumn{2}{c}{Encoded errors} & Chain error & Coverage \\
\cmidrule(lr){2-3}\cmidrule(lr){4-4}\cmidrule(lr){5-5}
Method & Count--PAD & Invalid & Activity & Sampling \\
 & & co-traveler & chaining & zeros \\
\midrule
Resample & 0 & 0 & .05 & 0 \\
IPF/IPU (train controls) & 0 & 0 & .06 & 0 \\
\midrule
BN & $\mathbf{0}$ & $\mathbf{0}$ & $\mathbf{0}$ & $38.4\pm6.0$ \\
VAE & $96.5\pm0.7$ & $4.5\pm0.7$ & $64.7\pm6.4$ & $\mathbf{57.6\pm4.9}$ \\
CTGAN & 100 & $14.0\pm0.3$ & $83.0\pm2.6$ & $\underline{53.9\pm6.6}$ \\
AR & $\mathbf{0}$ & $\mathbf{0}$ & $2.3\pm0.3$ & $34.7\pm4.1$ \\
DiaryBridge count-first & $\mathbf{0}$ & $\mathbf{0}$ & $\underline{0.4\pm0.1}$ & $26.9\pm7.6$ \\
DiaryBridge joint & $\underline{13.6\pm3.3}$ & $\underline{0.4\pm0.1}$ & $\underline{0.4\pm0.1}$ & $33.1\pm3.4$ \\
\bottomrule
\end{tabular}
\par\smallskip
\end{table}

We finally test whether generated diaries satisfy basic structural and activity-continuity requirements, and whether they reproduce household patterns absent from training but observed in the test data. We assess both properties using the unconditional samples from Section \ref{sec:res-fit}.

We first check whether the generated counts agree with the records they describe. Count–padding error is the percentage of households whose counts violate the encoding capacities or disagree with the occupied slots. For example, a household declaring two members should not contain person attributes in a third member slot. This measure identifies inconsistencies between the declared household structure and its encoded content.

We then check whether companion references identify admissible household members. Co-traveler error is the percentage of active trips with readable companion codes that name the traveler themself or a member outside the household roster. This check concerns who can be named as a companion; it does not require a matching trip in the companion’s diary, unlike the reciprocal-reference measures in Section \ref{sec:res-member-evidence}.

Finally, activity-chaining error measures discontinuities between successive trips by the same person. It is the percentage of adjacent trip pairs for which the first trip’s destination activity differs from the next trip’s origin activity. For example, a trip ending at work should be followed by one starting at work.

DiaryBridge count-first and AR have zero count–padding and co-traveler errors by construction. DiaryBridge joint, which does not enforce count–padding consistency, only produces count–padding errors in 13.6\% of households and invalid companion references in 0.4\% of eligible trips (Table \ref{tab:feasibility}). Both DiaryBridge samplers have an activity-chaining error of 0.4\%, compared with 2.3\% for AR, despite not enforcing activity continuity. BN has no activity-chaining errors because continuity is imposed during generation rather than learned.

We next ask whether a generator can reproduce household combinations missing from the training sample but present in the test data \citep{Kim2023FeasibleDiverse,Garrido2020RareCombinations}. A household pattern consists of its four household attributes (area type, home type, home ownership, and income) together with its vehicle and person counts. A pattern observed in test but absent from training is called a sampling zero: it is an unobserved training combination, not a forbidden combination or structural zero.

VAE and CTGAN have the highest coverage, but also produce padding errors in 96.5\% and 100\% of households. Their higher coverage therefore does not by itself establish greater ability to generate valid new diaries. BN, AR, and DiaryBridge all recover test-only patterns while maintaining low padding and co-traveler errors. 

\section{Conclusion}
\label{sec:conclusion}
We introduced DiaryBridge, a masked diffusion model that represents household attributes, vehicles, persons, and daily trips as a single linked record. The same model generates complete diaries or fills in missing information while preserving observed fields. Its count-first sampler generates the numbers of vehicles, persons, and trips before their attributes, ensuring agreement between counts and records. Experiments on the Regional Travel Survey demonstrate DiaryBridge’s ability to generate travel reflecting household context and recover shared trips using other persons’ diaries. The results also show improved aggregate demand accuracy over modern learned baselines from the population synthesis literature. 

However our evaluation has limitations. All experiments use one unweighted survey split, so they do not establish survey-weighted regional demand or transfer across surveys. Future work should extend the model to use local population statistics, represent multi-day diaries, and combine information from complementary surveys. 

\section*{Data and code availability}
The 2017/2018 Regional Travel Survey public files are distributed by the Metropolitan Washington Council of Governments \citep{TPB2021RTS}. The code is available at \url{https://github.com/PascalJD/diarybridge-release}.

\bibliographystyle{elsarticle-harv}
\bibliography{references}

\end{document}